# Image-based underwater 3D reconstruction for Cultural Heritage: from image collection to 3D. Critical steps and considerations

**Dimitrios Skarlatos** [1†]  **& Panagiotis Agrafiotis** [1,2†]

[1] Cyprus University of Technology, Civil Engineering and Geomatics Dept., Lab of Photogrammetric Vision 2-8 Saripolou str., 3036, Limassol, Cyprus; dimitrios.skarlatos@cut.ac.cy, panagiotis.agrafioti@cut.ac.cy

[2] National Technical University of Athens, School of Rural and Surveying Engineering, Department of Topography, Zografou Campus, 9 Heroon Polytechniou str., 15780, Athens, Greece; pagraf@central.ntua.gr

[†] Authors contributed equally to this work

**Abstract**   Underwater Cultural Heritage (CH) sites are widely spread; from ruins in coastlines up to shipwrecks in deep. The documentation and preservation of this heritage is an obligation of the mankind, dictated also by the international treaties like the Convention on the Protection of the Underwater Cultural Heritage which fosters the use of "non-destructive techniques and survey methods in preference over the recovery of objects". However, submerged CH lacks in protection and monitoring in regards to the land CH and nowadays recording and documenting, for digital preservation as well as dissemination through VR to wide public, is of most importance. At the same time, it is most difficult to document it, due to inherent restrictions posed by the environment. In order to create high detailed textured 3D models, optical sensors and photogrammetric techniques seems to be the best solution. This chapter discusses critical aspects of all phases of image based underwater 3D reconstruction process, from data acquisition and data preparation using colour restoration and colour enhancement algorithms to Structure from Motion (SfM) and Multi-View Stereo (MVS) techniques to produce an accurate, precise and complete 3D model for a number of applications.

## Introduction

Image-based underwater 3D reconstruction is a key tool for 3D record, map and model submerged heritage providing the 3D relief and the valuable visual information together. In this context, they represent an effective tool for research, documentation, monitoring and more recently public diffusion and awareness of UCH assets, through for example, virtual reality headsets, serious games, etc. [1][2][3]. They can serve also as a tool for the assessment of the state of preservation of the submerged heritage and its threats, natural or manmade. Depending on the archaeological needs and on the environmental conditions such as depth or water turbidity, sensors, techniques and methods may need to be used differently or may even be not suitable at all [4].

Despite the relative low cost of the image-based methods in relation to others, they present a major drawback; optical properties and illumination conditions of water severely affect underwater imagery and data acquisition process. Colours are lost as the depth increases, resulting in a green-blue image effect due to light absorption, which mainly influences red wavelength. Therefore, red channel histogram has fewer values compared to green and blue. Water also absorbs light energy and scatters optical rays creating blurred images, reducing the exploitable visibility to a few meters.

Moreover, refraction causes additional issues on the processing of the underwater imagery. In the literature, two different approaches are reported for dealing with the refraction effect, when the camera is completely submerged; the first one is based on the geometric interpretation for light propagation through various media (e.g. air – housing device – water) and the other on the application of suitable corrections, in order to compensate for the refraction. Some researchers use a pinhole camera for the estimation of the refraction parameters, while others calibrate the cameras with the help of an object of known dimensions, which is put underwater in situ [5]. Nowadays, self-calibration is widely applied for the camera-housing system, as it is assumed that refraction effects are compensated by the interior orientation parameters [6].

In this chapter, a brief reference to the state of the art in underwater 3D reconstruction of CH is followed by an analysis of the implications caused by the underwater environment to the Structure from Motion and Dense Image Matching techniques. Camera calibration, underwater network establishment and data acquisition issues are also discussed. Moreover, the need and use of image processing techniques in the underwater 3D reconstruction along with best practices, is discussed.

## State of the art in underwater image-based 3D reconstruction for Cultural Heritage

Underwater photogrammetry for seabed mapping has a long history, with initial systematic experiments dating back to the sixties [7]. With its versatility, low cost equipment and lately the high degree of automation as its main advantages, became the most widely used technique in underwater CH 3D reconstruction nowadays, with first report of underwater Structure from Motion application reported in [8]. Since shallow and deep waters impose different constraints, which influence the recording process, there are plenty applications reported in literature about underwater Cultural Heritage 3D reconstruction. In [4] a wide overview of the state of the art on the field is being reported. However, in the following paragraphs, some of the more interesting and recent works found in the literature are presented in respect to the data acquisition methodology; divers or robotics platforms.

### *Diver based data for underwater 3D reconstruction*

An early report mapping amphorae discovered in a sunken ship off the shore of Syria can be found in [9], where a digital orthophoto mosaic was generated. During the years that followed, sensors and cameras technology advancement together with the affordable waterproof housings and the availability of educational and low-cost commercial software for close range photogrammetry facilitated the spread of the image-based underwater 3D reconstruction.

During the recent years, a large number of published studies reports applications of photogrammetry and computer vision for underwater CH documentation for depths within the limits of most recreational diving certifications. Bruno et al. in [10], report over the documentation of an archaeological site in the Baiae underwater park, during experimental conservation operations. McCarthy and Benjamin in [11] discuss the 3D results from trials in Scotland and Denmark at depths of up to 30 m. Yamafune et al. in [12] present a methodology to record and reconstruct the wooden structures of a $19^{th}$ century shipwreck in southern Brazil and of a $16^{th}$ century shipwreck in Croatia. In [13] a case study of application of photogrammetric techniques for archaeological field documentation record in course of underwater excavations of the Phanagorian shipwreck is reported. In [14] a survey and 3D representation of two Roman shipwrecks using integrated surveying techniques for documentation of underwater sites is described. In [15] and [16] one of the first systematic approaches on the continuous 3D documentation of an underwater CH asset during the excavation process is reported. The site studied there is the classical shipwreck of Mazotos lying at 45m depth, thus beyond recreational diving certifications.

More recently, Abdelaziz and Elsayed [17] documented the archaeological site of the lighthouse of Alexandria situated at a varying depth of 2 to 9 metres. Until 2016, only 7200$m^2$ of the 13000 $m^2$ of the submerged site, were covered. Bruno et al. in [18] presented an interesting approach for studying and monitoring the preservation state of an underwater archaeological site, by combining the quantitative measurements coming from optical and acoustic surveys with the study of biological colonization and bio-erosion phenomena affecting ancient artefacts. In [19] some methods for underwater documentation were presented and their advantages and disadvantages reported. Authors in [20] reported on the complementarity between in situ studies and photogrammetry by presenting the feedback from a roman shipwreck in Caesarea, Israel.

Most of the aforementioned approaches use the commercially available software application Agisoft Photoscan© . There are few reports in the literature using open source software for such applications ([8][15][21]) where, for the reconstruction process, the Bundler software [22] was used to orient the images and retrieve the camera calibration parameters. The successive DIM step was then performed using the PMVS (Patch-based Multi-View Stereo) software [23]. For documenting a semi-submerged archaeological structure, Menna et al. in [24] adopted a photogrammetric method, initially developed for marine engineering applications. In this method, two separate photogrammetric surveys are carried out, one in air above the water level and one underwater. Then, through several special rigid targets, that are partially immersed, rigid transformations are computed to combine the two separate surveys in a unique reference system.

## *Autonomous Underwater Vehicle (AUV) and Remotely Operated Vehicle (ROV) based data for image-based underwater 3D reconstruction*

Image data acquisition carried out by scuba divers impose depth limit and several safety related constraints; limited bottom time and nitrogen narcosis (deeper than 30m) may lead to several consecutive dives by several divers for the same data acquisition campaign, thus increasing risks and complicating logistics. To overcome the aforementioned shortcomings, robotic platforms like AUVs and ROVs have been adopted, especially in cases where larger areas must be covered or access is dangerous and depth prohibiting. However, while in shallow waters the use of these platforms may be an option over divers, exceeding the recreational diving limits on depth and limiting bottom time, reduces choices. Provided that the ROV and AUV systems can perform data acquisition needed for image-based 3D reconstruction purposes, they seem to be a safe choice to reduce or completely abandon use of divers, especially in depths more than 30m where the use of artificial lighting is necessary. However, when it comes to real world applications, small ROVs are difficult to handle, especially in areas with strong currents, they are prone to water leaks and have limited operational time while larger and more reliable ROVs are expensive, requiring specialized boats and personnel for their operation, thus increasing the cost.

Captured data by these systems in image-based 3D reconstruction applications for CH consist mainly of optical data, but other sensors may also be on board, especially when larger ROVs and AUVs are employed. Typically, the imagery taken from an AUV or ROV system is acquired by following the principles of aerial photogrammetry. Captured data are processed using an SfM and MVS pipeline [25][26][27][28]. Additionally to the SfM MVS processing, the extracted feature points are matched and tracked into overlapping stereo-image pairs. This resulting information is then integrated with additional navigation sensor data such as a depth sensor, a velocity sensor and data from the Inertial Navigation System (INS) in order to implement a SLAM algorithm and compute the trajectory of the platform. This estimated trajectory and the 3D points resulting from the SfM-MVS processing are then used to reconstruct a global feature map of the underwater scene. This step is of high importance when it comes to submerged CH mapping since it enables the AUV or ROV pilot to be aware of the area covered and thus the completeness of the delivered 3D reconstruction.

In the literature, many studies describe similar approaches. Johnson-Roberson et al. in [29] adopted an AUV and a diver-controlled stereo imaging platform (for very shallow water) in order to document the submerged Bronze Age city at Pavlopetri, Greece. Bingham et al., in [30] developed techniques for large-area 3D reconstruction of a 4th c. B.C. shipwreck site off the Greek island of Chios in the north eastern Aegean Sea using an UAV. Mahon et al., in [26] presented a vision-based underwater mapping system for archaeological use in the same area. Another interesting approach is presented by Bosch et al. in [31] where an omnidirectional underwater camera mounted on an AUV was used for a mapping a shipwreck. In [32] an approach based on photogrammetry for surveying the Roman shipwreck Cap Bénat 4, at a depth of 328m using an ROV is presented. The Visual Odometry technique presented there provides real time results, sufficient for piloting the ROV from the surface vessel and ensures a millimetric precision on the final 3D results. Finally, in the work presented in [33], bathymetric maps of underwater archaeological sites in water depths between 50m and 400m and different turbidity conditions were generated using an ROV equipped with optical cameras, laser and a multibeam sonar. Expected results over the comparison of the three different sensors indicated that in every case the laser and multibeam results were consistent while in stereo imaging the point density was highly dependent on scene texture, which is high turbidity environments may render photogrammetric approaches useless.

The aforementioned studies highlight that underwater image-based 3D reconstruction is a tool that has been accepted and applied by many disciplines and experts. Even though this facilitates faster mapping of submerged cultural heritage, with impressive results, implementation of those techniques by non-experts or ignoring the difficulties addressed in this chapter, underlies the danger of producing non-accurate and unreliable results.

**Implications to bundle adjustment and Structure from Motion.**

Any given set of images of a specific object, captured from different viewpoints, must undergo bundle adjustment as part of the 3D reconstruction process. This task can be described as the simultaneous estimation of camera positions so that the bundles of rays from the images intersect in 3D spaces, both in common points and in control points, i.e, points with known coordinates in the reference system of choice. At any given photogrammetric project, the Bundle Adjustment (BA) is the critical task where all gross and systematic errors are to be detected, estimated and finally eliminated at a great extent. Remaining systematic errors will affect the final results, particularly the 3D reconstruction, in an unpredictable way. Moreover, these errors will remain undetectable, unless check points are utilised, as a mean to quantify the remaining errors. Camera's interior orientation is a potential systematic error source and as such, BA can be employed to resolve camera's (or cameras') geometry. This process is known as self-calibration.

Despite that BA was a well-established process in photogrammetry, the last decade this process is being replaced from Structure from Motion (SfM), which is a more generic process than BA and in fact includes robust BA as the last step. However, it differs significantly from conventional photogrammetry, where a priori knowledge for camera used, initial approximations of camera stations, and a set of control points is required. In fact, camera geometry and camera positions and orientation are solved automatically without the need to specify any a priori knowledge. These are estimated simultaneously using a highly redundant, iterative bundle adjustment procedure, based on a database of features automatically extracted from a set of multiple overlapping images [34]. The SfM approach is most suited to sets of images with a high degree of overlap that capture full three-dimensional structure of the scene viewed from many different positions, or as the name suggests, images derived from a moving sensor [35].

In essence SfM is more generic, as it includes both the automated task of feature points detection, descriptions and matching, followed by robust SBA [36]. Several variations exist, each one with its own characteristics, strengths and weaknesses [37]. The most critical task of the process is feature detection, description and matching, as blunders are unavoidable in this phase. Poor detection and matching might lead to incomplete alignment, erroneous alignment of few images or total failure of the alignment. In all cases, some blunders will remain to the final solution, even after robust SBA and will affect final 3D reconstruction. It is advised that these errors are manually, or semi automatically selected and removed during the alignment phase.

In a similar way, underwater 3D reconstruction employing SfM, enjoys speed, ease of use and versatility but suffers the same limitations and shortcomings. In fact, due to particularities of the environment, there are several reasons for the SfM to fail.

Many problems have been reported that tend to be particularly profound in the underwater environment, posing either hard limitations or shortcomings, which if properly addressed may be overcome. Shortcomings may be divided in two categories; environmental and computational, with the former ones need to be addressed during the acquisition phase and the latter ones being able to address during processing.

Environmental shortcomings affect acquisition process, including control point network establishment, stability and coordinate system definition. The deeper the site the more the shortcomings. Depth, increases the colour absorption, decreases light, reduces bottom time and enhances nitrogen narcosis effects. All these problems must be dealt during the acquisition phase, with proper planning and dive logistics. Some problems, such as camera calibration, colour aberration, vignetting etc, can be dealt at some extend with dome lens housings and prime camera lenses. Remaining environmental problems effects, may also be dealt computationally, provided they are not severe.

## Camera calibration

The obvious consideration on underwater photogrammetry is camera calibration, which although a trivial task in air, underwater implementation is not. Two media photogrammetry is governed by law's of physics, and therefore, collinearity equation may be modified and used for underwater camera calibration. Several authors have investigated the influence of flat or dome port in underwater photogrammetry, both in terms of geometry, and colour ([38][39][24]). The use of dome port, in theory completely removes refraction if the projection centre is positioned at the centre of the dome [39], but in general case, this is very difficult to achieve unless the camera and housing are being manufactured as a uniform body. Deviations below 1 cm from the concision of the two centres might be ignored, or absorbed by the central and tangential distortion parameters of the camera calibration model [40]. In case the camera is misaligned to the dome or a flat port is being used, the conventional distortion model will not suffice and a full physical model must be adopted [39] such as in [41] or [42], taking into consideration the glass thickness. Other researchers perform calibration on air and then compile parameters in underwater environment [43].
Chromatic aberration (Fig. 1), is severe in underwater environment and may result in several pixels deformation [24]. Although it seems as an irrelevant radiometric problem, it can affect image geometric properties during both calibration and 3D reconstruction, in several ways. For example, calibration channel (colour) and 3D reconstruction channel (colour), should be compatible (the same), otherwise deformations will occur. Therefore, good practice on underwater SfM, is selecting a channel to work with, both during alignment and 3D reconstruction phase, while for texturing the full colour images maybe used, instead of the single channel ones. Post processing to amend chromatic aberration could be applied but analysis of the actual image-point correction due to refraction shows that for close-range imaging, the actual aberration is depth dependent, and three dimensional problem ([39]4344).
Light absorption and vignetting affects are also significant, especially in wide angle lenses, but the they are bounded to radiometry without any geometric extension. Using underwater strobes is effective in small distances, and even then, the effect is not uniform (Fig. 1). Light absorption also affects clarity and crispness of images, hence deteriorating the performance of feature detectors. Backscattering, caused by floating particles and false strobe positioning, render photos useless by feature detectors (Fig. 2) Hence, these effects might influence texturing or orthoimage production. Underwater photography is a difficult task, governed by many issues, and must be mastered before performing photogrammetric documentation of a cultural site.

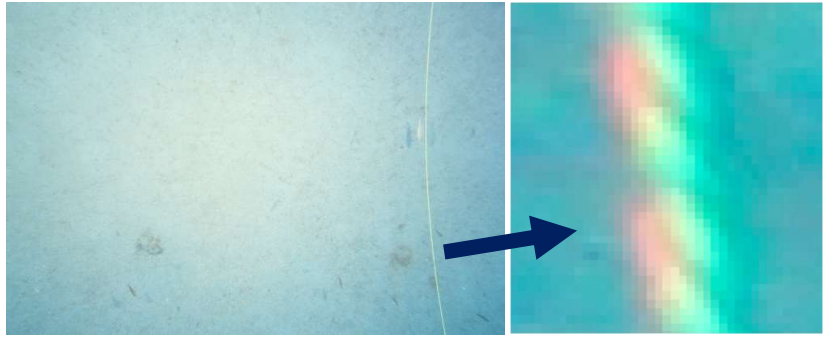

**Fig. 1.** Samples of underwater chromatic vignetting (left) and chromatic aberration in detail (right), where the colour shift of a white line is demonstrated.

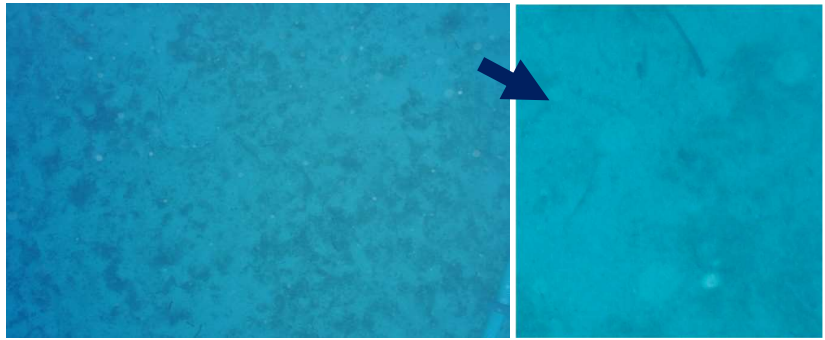

**Fig. 2.** Typical sample of backscattering effect, if lights are not positioned correctly. Such mild effects can be processed correctly from SfM, but when more profound, the results are unexpected.

### *Network of control points establishment and solving.*

Georeferencing of CH sites is a standard process in land sites, but underwater is a difficult task at depths more than 3m. Up to such depths, use of large poles allows surfacing of GPS receiver and correct geolocalization of a rather limited number of points, as the process is time consuming. In larger depths, use of buoys is not recommended as currents and waves do not allow vertical lines to the surface. Available systems for exact geolocalization of underwater sites, such as long baseline acoustic positioning system, may provide accuracy of up to few centimetres, but the cost of the system is so high that cannot be sustainable, only for archaeological purposes. Most sites are documented in local reference systems, as establishing a reference system is necessary if site is to be revisited for monitoring or due to multiple excavation periods.

Even so, establishing an underwater network of control points, is not a trivial task. Selecting position of control points (design), fixation of control points, measurement acquisition, become difficult to perform, the deeper the site is [6]. Limited bottom time, low visibility and poor communication underwater are challenging conditions, which render many land practices completely useless. The prevailing measuring methods in underwater CH documentation are tape measurements and photogrammetry, with the latter having a true advantage in terms of acquisition speed [6], as a whole site may be measured within a single dive, where tape measurements require several dives, complicating dive logistics and overall planning of the expedition.

Even so, computational aspects on solving the network should also be considered, since vertical reference in not given, like in land CH sites. Buoys suffer from currents and cannot provide true vertical reference, inverted hoses with air inside, may transfer depth from point to point and provide relative depth differences, but not absolute depth, and dive computers are accurate to 10cm and very unreliable as reading differs from day to day and from brand to brand. By using photogrammetry and a free network bundle adjustment, one may take advantage and relate several dive computer depth readings into a single solution and therefore provide vertical reference to a site. In a similar way when using only tape measurements for trilateration adjustment, depth readings remain unrelated measurements as the inherent unreliable vertical solution of trilateration, cannot take advantage of them in a holistic adjustment solution. In [6] authors, based on realistic assumptions, demonstrated that when using photogrammetric measurements for free network adjustment, to assign coordinates in the control points, the average $\sigma XY$ error is 0.02m and the average $\sigma Z$ error, is 0.02m. Similar values, when using trilateration and tape measurements, are $\sigma XY$ 0.06m and the average $\sigma Z$ error, is 0.64m. Nevertheless, they point out that different assumptions over network might change results, although photogrammetric measurements will always be more precise.

## Colour processing of underwater images

Despite the relative low cost of the image-based methods in relation to others, they present a major drawback in underwater environment; optical properties and illumination conditions of water severely affect underwater imagery. Colours are lost as the depth increases, resulting in a green-blue image due to light absorption, which affects mainly red wavelength. Therefore, red channel histogram has fewer values compared to green and blue. Water also absorbs light energy and scatters optical rays creating blurred images.

## *Caustics effect*

Even though the above phenomena affect RGB imagery in every depth, when it comes to shallow waters (less than 10m depth), caustics, the complex physical phenomena resulting from the projection of light rays being reflected or refracted by a curved surface (Fig. 3), seems to be the main factor degrading image quality for all passive optical sensors [46]. Unlike deep water photogrammetric approaches, where midday might be the best time for data capturing due to brighter illumination conditions, when it comes to shallow waters, the object to be surveyed needs strong artificial illumination, or images taken under overcast conditions, or with the sun low on the horizon, in order to avoid lighting artefacts on the seabed [46].

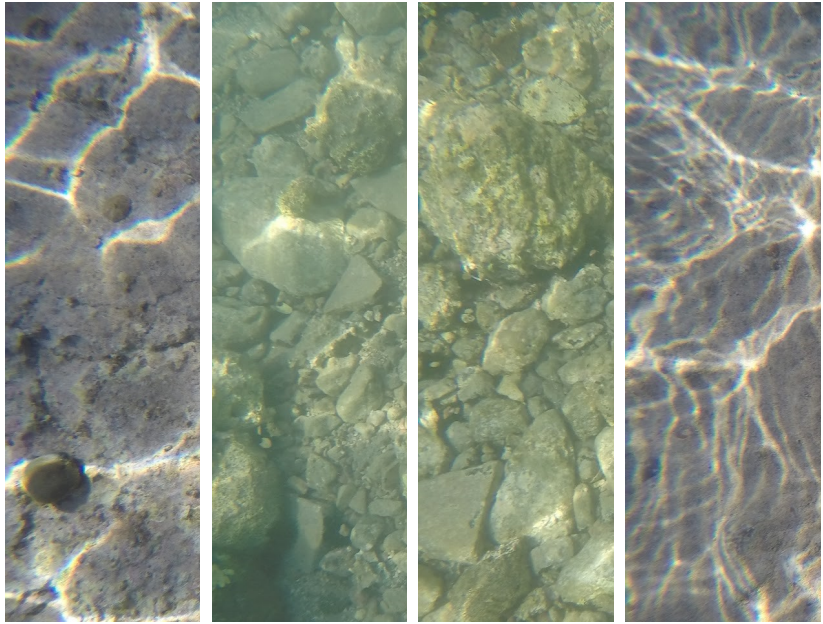

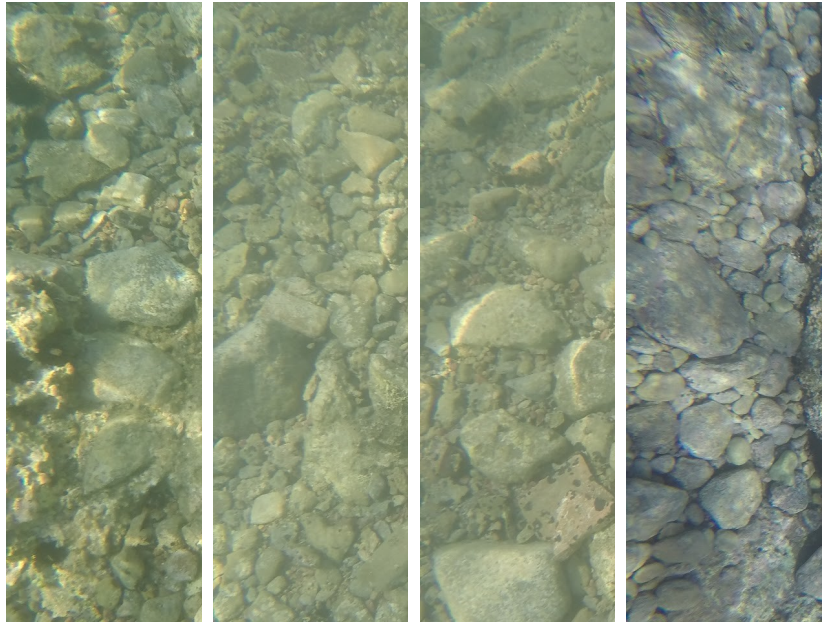

**Fig. 3.** Caustics of various patterns and density are present in the underwater imagery on shallow depths

If not avoided during the acquisition phase, caustics and illumination effects will affect image matching algorithms and are the main cause for dissimilarities in the generated textures and orthoimages, if these are the final results. In addition, caustics effects throw off most of the image matching algorithms, leading to less accurate matches [46].

In the literature, only a few techniques have been proposed for the removal of caustics from images and video in the context of image enhancement. Trabes et al., in [47] propose a technique which involves tuning a filter for sunlight-deflickering of dynamically changing underwater scenes. A different approach was proposed in [48] where a mathematical solution was presented involving the calculation of the temporal median between images within a sequence. The same authors later extend their work in [49] and propose an online sunflicker removal method which treats caustics as a dynamic texture. As reported in the paper this only works if the seabed or bottom surface is flat. In [50] authors propose a method based on analysing by a non-linear algorithm a number of consecutive frames in order to preserve consistent image components while filtering out fluctuations. Finally, Forbes et al., in [51] proposed a solution based on two small and easily trainable CNNs (Convolutional Neural Networks). This proposed solution was evaluated in terms of keypoint detection, image matching and 3D reconstruction performance in [46].

Despite the innovative and complex aforementioned techniques, addressing caustic effect removal with procedural methods requires that strong assumptions are made on the many varying parameters involved e.g. scene rigidity, camera motion, etc 45.

## *Underwater image restoration and underwater image enhancement*

During the last decades, the acquisition of correct or at least realistic as possible underwater colour imagery became a very challenging, as well as promising, research field which affects the image-based 3D reconstruction and mapping techniques [52]. To address these issues, two different approaches for underwater image processing are implemented according to their description in literature. The first one is image restoration. It is a strict method that is attempting to restore true colours and correct the image using suitable models, which parameterize adverse effects, such as contrast degradation and backscattering, using image formation process and environmental factors, with respect to depth ([53][54][55][56][57]). The second one uses image enhancement techniques that are based on qualitative criteria, such as contrast and histogram matching [58][59]. Image enhancement techniques do not consider the image formation process and do not require environmental factors to be known a priori [52][60]. In both approaches, recent advances in machine and deep learning facilitated the implementation of new improved techniques [61] for underwater image processing however, due to the lack of sufficient and effective training data, the performance of deep learning-based underwater image enhancement algorithms do not match in many cases the success of recent deep learning-based high-level and low-level vision problems [61].

## *Pre-processing or post-processing the underwater imagery*

Having developed various underwater image colour restoration and colour enhancement techniques, experts in underwater image-based 3D reconstruction faced the challenge of exploiting them and integrate them into the reconstruction steps. This integration is usually tackled with two different approaches; the first one focuses on the enhancement of the original underwater imagery before the 3D reconstruction in order to restore the underwater images and potentially improve the quality of the generated 3D point cloud. This approach in some cases of non-turbid water [52][60] proved to be unnecessary and time-consuming, while in high-turbidity water it seems to have been effective enough [62]. The second approach suggests that, in good visibility conditions, the colour correction of the produced textures or orthoimages is sufficient and time efficient [52][60].

Recently, a combination of the above was proposed in [63]. There, an investigation as to whether and how the pre-processing of the underwater imagery using five implemented image enhancement algorithms affects the 3D reconstruction using automated SfM-MVS software is performed. This work follows and completes the work of presented in [52] and [60].

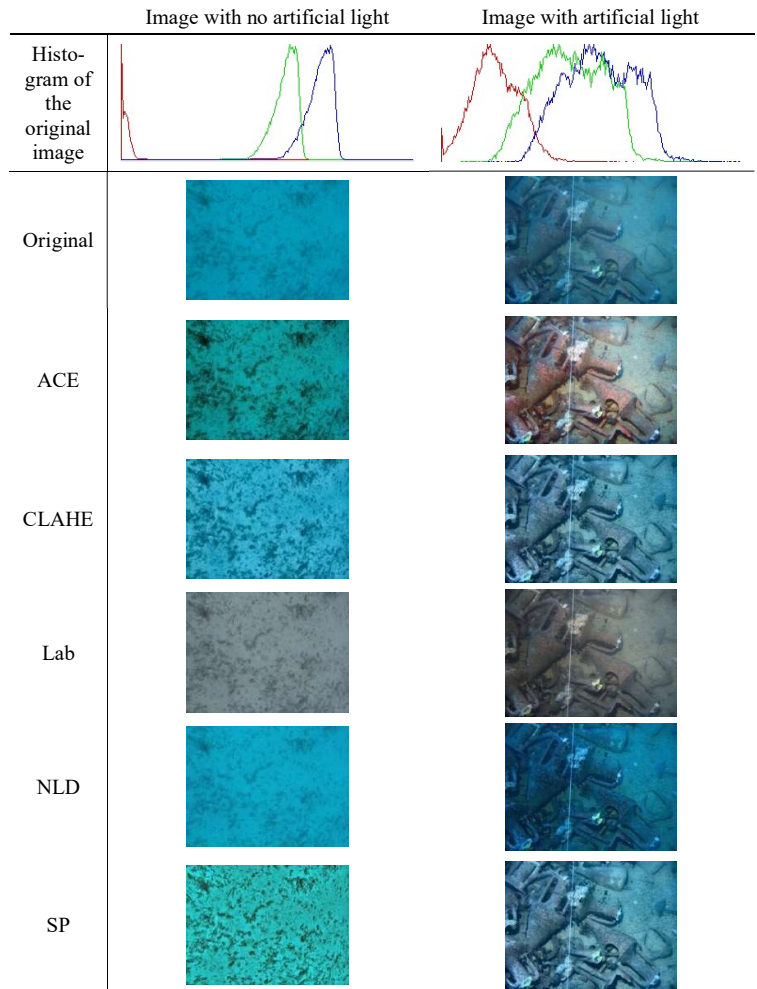

**Fig. 4.** Example images without artificial light (left column) at a depth of 34.5m captured by an ROV and with artificial light (right column) at a depth of 45m (Credits: Photogrammetric Vision Lab. of Cyprus University of Technology for the left column and MARELab, University of Cyprus, for the images of the right column)

Specifically, each one of the presented algorithms in this article is evaluated according to its performance in improving the results of the 3D reconstruction using specific metrics over the reconstructed scenes of the five different datasets of submerged Cultural Heritage. To this end underwater imagery ensuring different environmental conditions (i.e., turbidity etc.), depth, and complexity was used. Results suggest that the 3D reconstructions were not significantly improved by the applied methods, probably the minor improvement obtainable with the LAB colour enhancement algorithm [64] could not justify the effort to pre-process hundreds or thousands of images are required for larger models.

In the case of an underwater 3D reconstruction, the tool presented in [63] can be employed to try different combinations of methods and quickly verify if the reconstruction process can be improved somehow. However, as can be observed in Fig. 4, if no artificial light is present from a depth and below, images cannot be improved due to severe lack of the red channel, and most of the image enhancement methods fail.

A strategy that is suggested is to pre-process the images with the LAB [64] method trying to produce a more accurate and dense 3D reconstruction and, afterwards, to enhance the original images with another method such as ACE [65] to achieve a textured model more faithful to reality. Employing this tool for the enhancement of the underwater images ensures to minimize the pre-processing effort and enables the underwater community to quickly verify the performance of the different methods on their own datasets.

## Conclusions

This chapter discussed critical aspects of all phases of image-based underwater 3D reconstruction process, from data acquisition and data preparation using image processing techniques to Structure from Motion (SfM) and Multi-View Stereo (MVS) techniques to produce an accurate, precise and complete 3D representation of the submerged heritage for a number of applications. It is straightforward that image-based 3D modelling of CH underwater sites offers the best performance to cost ratio. It is affordable, easy and fast, while offers excellent 3D spatial resolution and important visual information. However, it heavily depends on visibility, which renders the method inadequate for turbid waters. Quality of final results depend on many factors and are highly variable, depending on environmental conditions and data acquisition experience. The most important of these parameters is the camera to object distance reducing the field of view of a single image, minimizing the distance from the object and rendering full object coverage a challenge for any diver or ROV operator. Therefore, current bottleneck of what seems a flawless 3D reconstruction and texturing technique for VR applications, are illumination problems colour variations, processing power imitations, experience over data acquisition and reference system definition.

## Acknowledgements

Part of the work presented here conducted in the context of the iMARECULTURE project (Advanced VR, iMmersive Serious Games and Augmented REality asn-Tools to Raise Awareness and Access to European Underwater CULTURal heritagE, Digital Heritage) that has received funding from the European Union's Horizon 2020 research and innovation programme under grant agreement No 727153. Authors would like also to thank M.A.RE Lab from University of Cyprus and the lead archaeologist Prof. S. Demesticha for providing data from several underwater sites, and moreover challenging the authors to overcome problems and shortcomings of the 3D documentation process in underwater CH.